\documentclass{article}
\usepackage{spconf,amsmath,graphicx}
\usepackage{algorithm}
\usepackage{algorithmic}
\usepackage{bm}
\usepackage{amsfonts}
\title{A RANDOM GOSSIP BMUF PROCESS FOR NEURAL LANGUAGE MODELING}
\name{Yiheng Huang$^{\star}$\quad Jinchuan Tian$^{\star}$\quad Lei Han$^{\star}$\quad Guangsen Wang$^{\star}$\quad Xingchen Song$^{\dagger\dagger}$\quad Dan Su$^{\star}$\quad Dong Yu$^{\dagger}$ \thanks{The first two authors contribute equally. Thanks to Yuexian Zou, Zheng-Hua Tan, Mingzhi Lin and Haidong Rong for useful discussions. © 2020 IEEE.  Personal use of this material is permitted.  Permission from IEEE must be obtained for all other uses, in any current or future media, including reprinting/republishing this material for advertising or promotional purposes, creating new collective works, for resale or redistribution to servers or lists, or reuse of any copyrighted component of this work in other works.}}

\address{{$^{\star}$} Tencent AI Lab  \\ 
                {$^{\dagger\dagger}$} Department of Computer Science and Technology, Tsinghua University, Beijing, China\\
                {${\dagger}$} Tencent AI Lab, Bellevue, WA, USA}
\begin{document}
%\ninept
%
\maketitle

\newtheorem{definition}{Definition}
\newtheorem{theorem}{Theorem}
\newtheorem{proof}{Proof}

\begin{abstract}
Neural network language model (NNLM) is an essential component of industrial ASR systems. One important challenge of training an NNLM is to leverage between scaling the learning process and handling big data. Conventional approaches such as block momentum provides a blockwise model update filtering (BMUF) process and achieves almost linear speedups with no performance degradation for speech recognition. However, it needs to calculate the model average from all computing nodes (e.g., GPUs) and when the number of computing nodes is large, the learning suffers from the severe communication latency. As a consequence, BMUF is not suitable under restricted network conditions. In this paper, we present a decentralized BMUF process, in which the model is split into different components, each of which is updated by communicating to some randomly chosen neighbor nodes with the same component, followed by a BMUF-like process. We apply this method to several LSTM language modeling tasks. Experimental results show that our approach achieves consistently better performance than conventional BMUF. In particular, we obtain a lower perplexity than the single-GPU baseline on the wiki-text-103 benchmark using 4 GPUs. In addition, no performance degradation is observed when scaling to 8 and 16 GPUs.
\end{abstract}
\section{Introduction}
\label{sec:intro}

Machine learning, and in particular deep learning technology \cite{DeepLearning} powers many aspects of modern lives. At the core of deep learning lies the deep neural networks (DNNs), long short term memory networks (LSTMs), transformers, convolutional neural networks (CNNs) and their variants. These technologies have been widely implemented in a plenty of fields, such as language modeling \cite{Transformerxl,jozefowicz2016exploring}, natural language processing (NLP) \cite{vaswani2017attention,devlin2018bert} and large vocabulary continuous speech recognition (LVCSR) \cite{dahl2011context,graves2014towards}. As data size and model complexity increase, one essential challenge is to leverage between scaling the learning procedure and handling big data. Usually, to train a neural language model on nowadays language modeling datasets with competitive accuracy requires a high-performance computing cluster.  

Many works have been proposed to scale up the capability of deep learning. For example, DistBelief \cite{dean2012large} utilizes thousands of machines to train various deep machines with an asynchronous SGD (ASGD) procedure called \textit{Downpour SGD}. Hogwild \cite{recht2011hogwild} employs a lock-free ASGD procedure which is suitable for sparse gradients. Elastic averaging SGD \cite{zhang2015deep} has been proposed  recently and is the state-of-art asynchronous parameter-server method. We refer to \cite{ben2018demystifying} as a good survey for introducing these algorithms.  

Another popular approach to introduce parallelism is to average models (MA) directly \cite{Zinkevich2010,Miao2014}. MA updates local models independently on each worker and average them only once \cite{Zinkevich2010} or every a few iterations \cite{Miao2014}. These methods achieve nearly linear speedups but suffer from accuracy degradation \cite{Block-Momentum,povey2014parallel}. BMUF \cite{Block-Momentum}, proposed to tackle the degradation problem in MA, is widely used in speech recognition \cite{li2018improving,zhaoyou2019}. BMUF introduces a blockwise model update filtering process to stabilize the training. BMUF outperforms the traditional model averaging (MA) as well as alternating direction method of multipliers (ADMM) while enjoying the advantage
of low communication costs of such methods. Chen and Huo \cite{Block-Momentum} reported a performance with $28$X speedup with $32$ GPUs while also achieve better accuracy than the single-GPU SGD on large vocabulary speech recognition tasks. The results in \cite{Empirical17} showed that BMUF outperforms EASGD and ASGD on speech recognition tasks. 

All of the above methods utilize a centralized parameter-server structure for communications or use an \textit{all-reduce} process to average local models. Issues occur if one of these training nodes gets stuck which makes the whole training process hang up. Worse still, when the number of nodes is large, the problem of communication latency should be carefully considered. In this paper, we extend the BMUF process to a decentralized network topology. Each node of this training network only needs to communicate with a small number of its neighbors. During training, the entire model is split into multiple components, and each node randomly selects a few neighbor nodes to communicate. We refer this process to as \textit{gossip} \cite{jin2016scale} . Then this node aggregates these components from selected neighbors, and then performs a BMUF-like process as we mentioned above. In the experiments, we evaluate the proposed method on two benchmark datasets (e.g., wiki-text-103 and Gutenberg). The results show that our method consistently outperforms conventional approaches using $4$, $8$ and $16$ GPUs, with higher accuracy, lower variance, and comparable speedups.

%to \\\texttt{icassp2020@cmsworkshops.com}.

\section{Algorithms}
\label{sec:method}

%All printed material, including text, illustrations, and charts, must be kept
%within a print area of 7 inches (178 mm) wide by 9 inches (229 mm) high. Do
%not write or print anything outside the print area. The top margin must be 1
%inch (25 mm), except for the title page, and the left margin must be 0.75 inch
%(19 mm).  All {\it text} must be in a two-column format. Columns are to be 3.39
%inches (86 mm) wide, with a 0.24 inch (6 mm) space between them. Text must be
%fully justified.
In this section, we introduce the detailed training procedure. This general approach can be applied to train any type of deep  machines such as DNNs, LSTMs, CNNs, GRUs and Transformers, etc. 

\subsection{Network Topology}
\label{sec:topology}
\begin{figure}[htb]
\begin{minipage}[b]{.48\linewidth}
  \centering
  \centerline{\includegraphics[width=4.0cm]{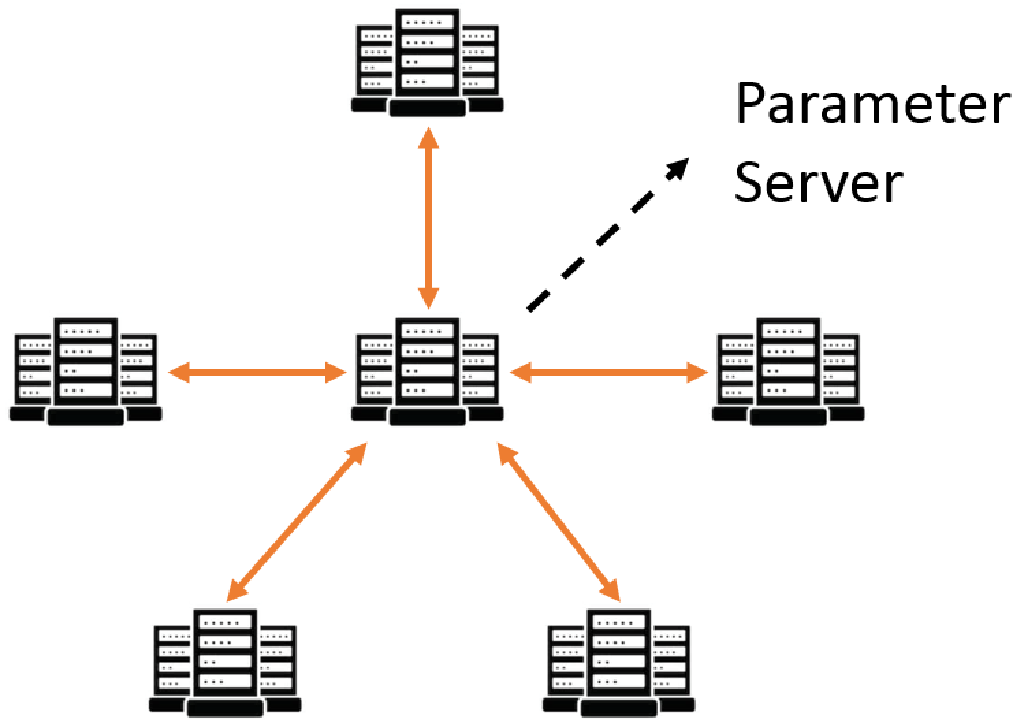}}
  %\vspace{1.5cm}
  \centerline{(a) Centralized Topology}\medskip
  \end{minipage}
\hfill
\begin{minipage}[b]{.48\linewidth}
  \centering
  \centerline{\includegraphics[width=3.8cm]{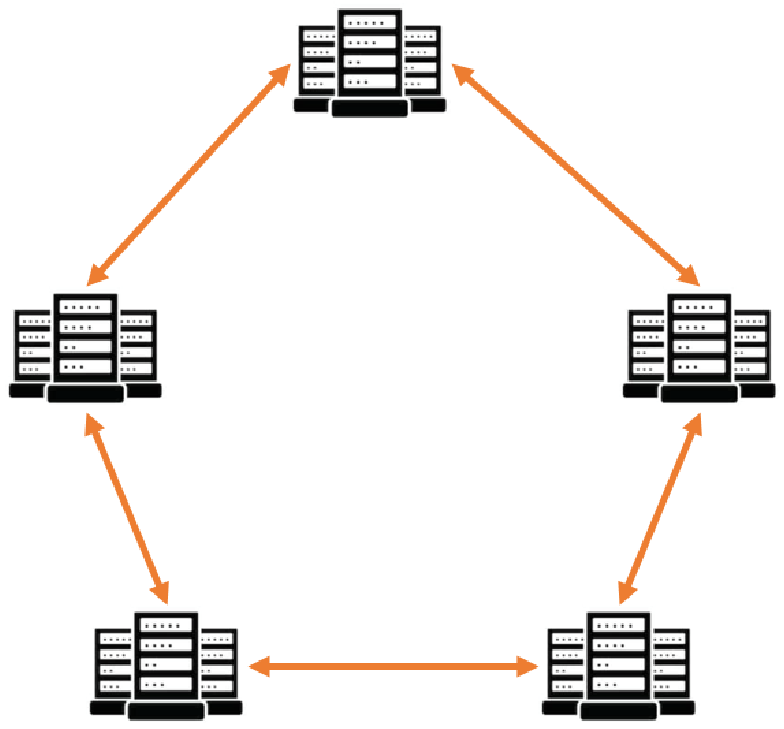}}
  %\vspace{1.5cm}
  \centerline{(b) Decentralized Topology}\medskip
  \end{minipage}

\caption{Illustrations of network topology}
\label{fig:topology}
\end{figure}
The topology of our network is decentralized as shown in figure \ref{fig:topology} (b). The formal description of our topology is defined in definition \ref{def:topology}.
\begin{definition} 
Assume there are $n$ nodes in total, numbered as $0,1,...,n-1$. We say the network forms a \emph{$k$-symmetric ring topology}, if node $i$ is connected to nodes $ (i-1)\% n, ..., (i-k)\% n, (i+1)\% n, ..., (i+k)\%n$, $ \forall 1\leq i \leq n$.
\label{def:topology}
\end{definition}
In the topology defined in definition \ref{def:topology}, each node connected to the $i$-th node is called the \textit{neighbor} of node $i$, and $k$ is referred to as the \textit{symmetric-degree} of this topology. 
%Assume $\mathbf{D}$ to be the full training data, the training data is evenly partitioned into $T$ non-overlapping blocks, and each block is partitioned into $N$ non-overlapping splits, this partition can be formally presented as follows:
% \begin{align}
%\mathbf{D}=\{\mathbf{D}_i|i=1,...,T\} \nonumber\\
%\mathbf{D}_i = \{\mathbf{D}_{ij}|j=1,...,N\}\\
%\forall i,j,k,l   \text{ }  \mathbf{D}_{ij} \cap \mathbf{D}_{kl} = \emptyset \nonumber
%\end{align}

\subsection{Details of Algorithms}
\label{sec:algorithm}
In this section, we give the details of our implementation. In our algorithm, we assume the network is formed with a \textit{$p$-symmetric ring topology} as defined in definition \ref{def:topology}, where $p$ is the \textit{symmetric-degree}. The model parameter $\bm{\theta}$ is split into $m$ components as $\bm{\theta} = [\bm{\theta}_1^T,...,\bm{\theta}_m^T]^T$, and the training data $D$ is evenly split into $n$ splits $D = \cup D^k$, where $n$ is the total number of nodes.   
\begin{algorithm}[H]
\caption{gossip-MA (gossip-BMUF) for node $k$}
\label{gossip-BMUF}
\begin{algorithmic}[1]
\REQUIRE initial model $\bm{\theta}_0$
\REQUIRE components of the model $\bm{\theta} = [\bm{\theta}_1^T,...,\bm{\theta}_m^T]^T$ 
\REQUIRE slots $\bm{\Delta}^k, \bm{\omega}^k, \bm{G}^k$ with the same shape as $\bm{\theta}$.
\REQUIRE training data with labels $D$
\REQUIRE synchronous period $H_i$ for each component $\bm{\theta}_i$
\REQUIRE number of gossip neighbors $q\leq 2p$
\REQUIRE momentum $\eta$ and block learning rate $\zeta$
\REQUIRE learning rate $\alpha_t$
\STATE $\bm{\theta}_{0}^{k}\leftarrow\bm{\theta}_0, \bm{\omega}^{k}\leftarrow\bm{\theta}_0, \bm{\Delta}^k \leftarrow \bm{0}, \bm{G}^k \leftarrow \bm{0}$
\FOR{$t = 1,...,T$}
  \STATE sample a mini-batch from $D^k_t$  and calculate the gradients $\bm{g}_{t}^{k}$ from this mini-batch
  \FOR{$i = 1,...,m$ \textbf{parallel}}
    \STATE $\bm{\theta}_{i,t}^{k} \leftarrow \bm{\theta}_{i,t-1}^{k} - \alpha_t\cdot\bm{g}_{i,t}^{k}$
    \IF{$t \bmod H_i == 0$}
    \STATE randomly choose $q$ neighbors $k_1,...,k_q$
    \STATE $\bar{\bm{\theta}}_{i,t}^{k} \leftarrow \frac{1}{q+1}\left(\bm{\theta}_{i,t}^{k} + \sum_{1\leq j \leq q}\bm{\theta}_{i,t}^{k_j}\right)$
    \IF{gossip-BMUF} 
     \STATE $\bm{G}_{i}^k \leftarrow \bar{\bm{\theta}}_{i,t}^{k} - \bm{\theta}_{i,t-H_i}^{k}$
     \STATE $\bm{\Delta}_{i}^k \leftarrow\eta\bm{\Delta}_{i}^k + \zeta \bm{G}_{i}^k$
     \STATE $\bm{\omega}_{i}^{k} \leftarrow \bm{\omega}_{i}^{k} + \bm{\Delta}_{i}^k$
     \STATE $\bm{\theta}_{i,t}^{k} \leftarrow \bm{\omega}_{i}^{k} + \eta\bm{\Delta}_{i}^k$
     \ELSIF{gossip-MA}
      \STATE $\bm{\theta}_{i,t}^{k} \leftarrow \bar{\bm{\theta}}_{i,t}^{k}$
    \ENDIF
    \ENDIF
  \ENDFOR
\ENDFOR
\RETURN $\bm{\theta} = \frac{1}{n}\sum_{k=1}^{n}\bm{\theta}_{T}^{k}$
\end{algorithmic}
\end{algorithm}
 In algorithm \ref{gossip-BMUF}, when $q=2p$, \textit{gossip-MA} and \textit{gossip-BMUF} are degenerated to two algorithms which we name them as \textit{local-MA} and \textit{local-BMUF} respectively. In \textit{local-MA} and \textit{local-BMUF}, since all of the neighbor nodes are included in the average process, no \textit{gossip} factor is introduced. 

Specifically, two important aspects in algorithm \ref{gossip-BMUF} need to be emphasized. First, different components of the model could have different synchronous periods. Second, different components may randomly choose different neighbors to communicate in each gossip process.

\subsection{Statistical Analysis}
In this section, we investigate the statistical properties of the proposed MA method (see algorithm \ref{simEMA}). The result shows that under mild assumptions, the MA estimator will asymptotically approach the optimal but with a bias term that is proportional to the number of workers. Similar analyses can be derived for the gossip algorithms which we will leave to the future work.  Some basic concepts according to convexity can be found in \cite{nesterov2018lectures}. In the following, assume $\bm{\theta}^k$ is the parameter for worker $k$ with dimension $d$. $\bm{\theta}_t = [\left(\bm{\theta}_{t}^1\right)^{T},...,\left(\bm{\theta}_{t}^n\right)^{T}]^T$ is the concatenated parameter vector of worker $1$ to $n$ at step $t$. $\bm{\theta}_*$ is the optimal parameter to be estimated, and $\bm{\theta}_*\mathbf{1}=[\bm{\theta}_{*}^{T},...,\bm{\theta}_{*}^{T}]^T$ is the vector of $\bm{\theta}_*$ replicated $n$ times. 
\label{sec:analysis}
\begin{algorithm}[H]
\caption{Simple MA}
\label{simEMA}
\begin{algorithmic}[1]
\STATE initialize each worker $i$, $\bm{\theta}_{0}^i\leftarrow\bm{\theta}_0$
\FOR{$t\in \{0,...,T\}$}
  \STATE $\bar{\bm{\theta}}_{t} = \frac{1}{n}\sum_{j=1}^{n}\bm{\theta}_{t}^j$ 
  \STATE $\bm{\theta}_{t+1}^i = \bar{\bm{\theta}}_{t} - \alpha_t \nabla f_i(\bar{\bm{\theta}}_{t};X_{t,i})$
\ENDFOR
\end{algorithmic}
\end{algorithm}
where $\alpha_t$ is the learning rate at step $t$ and $X_{t,i}$ is the data feed to worker $i$ at step $t$.

\begin{theorem}
\label{theorem1}
Let f be a m-strongly convex function with L-Lipschitz gradients. Assume that we can sample gradients g = $\bigtriangledown f(\bm{\theta};X_i)+\xi_i$. with additive noise with zero mean $\mathbb{E}[\bm{\xi}_i]=0$ and bounded variance $\mathbb{E}[\bm{\xi}_i^T\bm{\xi}_i] \leq \sigma^2$. Then, running the simple MA algorithm, with constant step size $0< \alpha \leq \frac{2}{m+L}$, the expected sum of squares convergence of the local parameters to the optimal $\theta_*$ is bounded by 

\begin{equation}
\mathbb{E}\left[|\bm{\theta}_t\!\! -\!\!\bm{\theta}_*\bm{1}|^2\right]  \leq   \left( 1\!\!-\!\! 2\alpha\cdot\frac{mL}{m+L} \right)^t
|\bm{\theta}_0-\bm{\theta}_*\bm{1}|^2+ n\frac{m+L}{2mL}\alpha\sigma^2
\end{equation}

\proof. Following a similar proof pipeline, using the inequalities (14)-(23) from the appendix of \cite{jin2016scale}, we obtain
\begin{align}
\label{proofe1}
    \mathbb{E}\left[|\bm{\theta}_{t+1}-\bm{\theta}_*\bm{1}|^2|\bm{\theta}_t\right]\leq \left(1-2\alpha_t\frac{mL}{m+L}\right)\cdot \nonumber \\ 
    n\left(\bar{\bm{\theta}}_{t}-\bm{\theta}_*\right)^T\left(\bar{\bm{\theta}}_{t}-\bm{\theta}_*\right) + n\alpha_t^2\sigma^2
\end{align}

The term  \[(\bar{\bm{\theta}}_t-\bm{\theta}_*)^T(\bar{\bm{\theta}}_t-\bm{\theta}_*)\] in equation (\ref{proofe1}) can be written as
\begin{equation}
    \sum_{j=1}^{d}\left(\frac{1}{n}\sum_{i=1}^n\left(\theta_{t}^i\right)^j-\theta_*^j\right)^2 = \sum_{j=1}^{d}\left(\sum_{i=1}^n\frac{1}{n}\left(\left(\theta_{t}^i\right)^j-\theta_*^j\right)\right)^2
\end{equation}
where $d$ is the dimension of vector $\bm{\theta}_{t}^i$. Using the Cauchy-Schwarz inequality, we have
\begin{align}
\label{inequale2}
 & \sum_{j=1}^{d}(\frac{1}{n}\sum_{i=1}^n(\left(\theta_{t}^i\right)^j-\theta_*^j))^2 \leq
   \sum_{j=1}^{d}( \frac{1}{n}\cdot(\sum_{i=1}^n(\left(\theta_{t}^i\right)^j-\theta_*^j)^2)) \nonumber\\
 & = \frac{1}{n}(\sum_{i=1}^n|\bm{\theta}_{t}^i-\bm{\theta}_*|^2)= \frac{1}{n}\cdot|\bm{\theta}_t-\bm{\theta}_*\bm{1}|^2
\end{align}
then, by substituting (\ref{inequale2}) to (\ref{proofe1}) we obtain 

\begin{align}
\label{algin2}
\mathbb{E}\left[|\bm{\theta}_{t+1}-\bm{\theta}_*\bm{1}|^2\mid\bm{\theta}_t \right] \leq \left(1-2\alpha_t\frac{mL}{m+L}\right)\cdot \nonumber\\
|\bm{\theta}_t-\bm{\theta}_*\bm{1}|^2 + n\alpha_{t}^2\sigma^2
\end{align}
assume $\alpha_t=\alpha$, we derived 
\begin{align}
\mathbb{E} \left[|\bm{\theta}_{t+1}-\bm{\theta}_*\bm{1}|^2\right] = \mathbb{E}_{\bm{\theta}_t}\mathbb{E}\left[|\bm{\theta}_{t+1}-\bm{\theta}_*\bm{1}|^2\mid\bm{\theta}_t \right]\nonumber \\ \leq \left(1-2\alpha\frac{mL}{m+L}\right)\cdot
\mathbb{E}\left[|\bm{\theta}_t-\bm{\theta}_*\bm{1}|^2\right] + n\alpha^2\sigma^2 \leq \nonumber\\
\left(1-2\alpha\frac{mL}{m+L}\right)^{t+1}\cdot|\bm{\theta}_0-\bm{\theta}_*\bm{1}|^2 + n\frac{m+L}{2mL}\alpha\sigma^2
\end{align}
which completes the proof.
\end{theorem}

Theorem \ref{theorem1} indicates that the upper bound of the MA estimator introduces a bias term proportional to the number of workers, which is consistent with the experimental results.

\section{EXPERIMENTS AND RESULTS}
\label{experiments}
Experiments are implemented on MPI-based HPC machine learning platform which contains $16$ nodes on two machines. Each node is an Nvidia Telsa M40 GPU with memory of 24GB. A 10 Gbps bandwidth network is configured to connect these two machines. All experiments are implemented using TensorFlow equipped with Horovod \cite{sergeev2018horovod}. 
\subsection{Datasets Description}
We choose two language model benchmark tasks to conduct the evaluations. The first dataset is wiki-text-103 \cite{merity2016pointer}, which contains $0.1$B tokens for training with OOVs replaced by $<$unk$>$. The size of the vocabulary is $267735$. The second dataset is Gutenberg\footnote{http://www.openslr.org/resources/12/original-books.tar.gz}. We use $95\%$ data for training and $5\%$ for testing. The training corpus contains $0.13$B tokens with OOVs replaced by $<$unk$>$, and the vocabulary is truncated at a frequency of $7$ and has $280811$ words in total.   
\subsection{Implementation Details}
Standard LSTM structure \cite{Graves2013} is used as a basis. In order to reduce the computations, an LSTMP \cite{LSTMP} layer with 2048 hidden units and a projection of 512 units is chosen. The word embedding dimension is 512. Each mini-batch contains $5120$ tokens. In wiki-text-103, a truncated BPTT for $40$ steps are used during backward pass, while for Gutenberg we use $20$ steps instead. We use dropout with a keep probability of 0.9 after the embedding layer and LSTMP layer, the bias of the LSTM forget gate were initialized to 1.0. To handle the large vocabulary problem, sampling methods such as CANE \cite{cane} can be used to speed up the training progress. Here we use \textit{adaptive-softmax} \cite{adaptive-softmax} with a tail projection factor $2$. The cluster number is chosen as $6$, and the head dimension is chosen to be 8192. And the tail clusters are split by averaging over the word frequencies. Adagrad \cite{adagrad} is chosen to be the optimizer. In all experiments, the epochs are set to be $20$. The initial learning rate is set to be $0.1$ with an exponential decay rate of $0.9$ after every epoch. The norm of gradients for LSTM is clipped by 10.0. For the parameters related to BMUF, the block learning rate is set to $1.0$ and the momentum is set to $0.9$ in all experiments. The model is split as follows: (1) embeddings are evenly split into $8$ shards; (2) LSTM weights form one group; (3) LSTMP weights form another group; (4) Head weights in \textit{adaptive-softmax} form one group; (5) Weights of each tails in \textit{adaptive-softmax} form their individual groups.    

\subsection{Experimental Results}
We compare our algorithm with Block-Momentum \cite{Block-Momentum} and MA \cite{Zinkevich2010,Miao2014}. In algorithm \ref{gossip-BMUF}, the sync-periods $H_i$ need to be pre-determined for each component of the model. Here we test two settings of sync-periods, where the first one is set to $(8)$ indicating that we synchronize all components every $8$ mini-batches. The second one is $(16,128)$, which indicates that we synchronize each embedding shards every $128$ mini-batches and synchronize other components every $16$ mini-batches. We mainly report the results of the setting $(16,128)$, since the accuracy of the setting $(16,128)$ is slightly better than $(8)$ with much faster training speed. In $4$-GPUs experiments, the parameters $p$ and $q$ in algorithm \textit{gossip-MA} and \textit{gossip-BMUF} is set to be $1$ and $1$ respectively, and they are set to $1$ and $2$ in \textit{local-MA} and \textit{local-BMUF}. In $8$-GPUs experiments, $p$ and $q$ are set to $2$ and $2$ in \textit{gossip-\{MA,BMUF\}} while in \textit{local-\{MA,BMUF\}} $p$ and $q$ are set to $2$ and $4$. Finally, in $16$-GPUs experiments, $p$ is set to $3$ and $q$ is set to $2$ in \textit{gossip-\{MA,BMUF\}} and $6$ in \textit{local-\{MA,BMUF\}}. In the above settings, we choose $p$ empirically according to $p=log_2(n)-1$, where $n$ is the number of workers. 

\begin{figure}[htb]

\begin{minipage}[b]{.48\linewidth}
  \centering
  \centerline{\includegraphics[width=4.8cm]{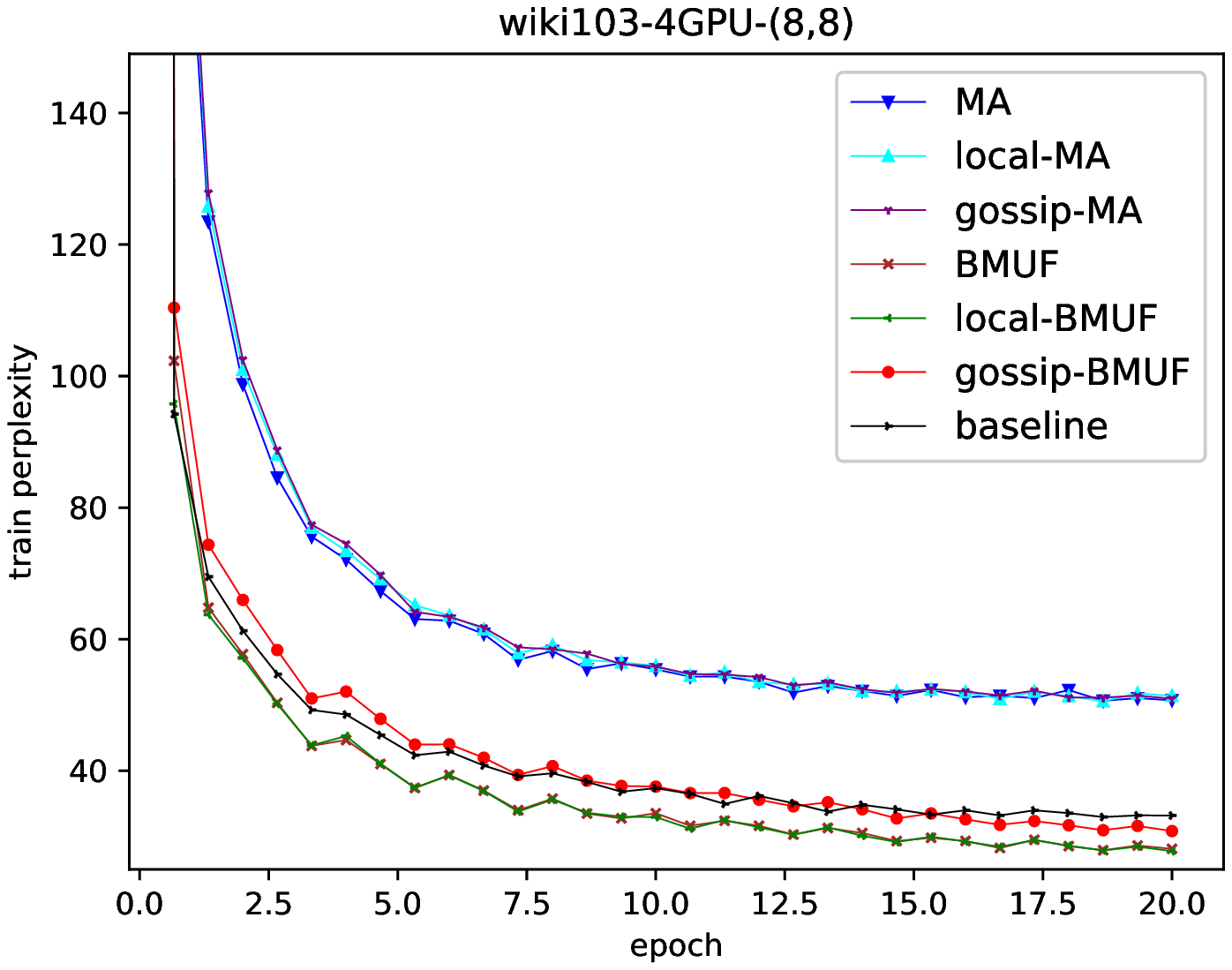}}
  %\vspace{1.5cm}
  \centerline{(a) sync-period 8}\medskip
  \end{minipage}
\hfill
\begin{minipage}[b]{0.48\linewidth}
  \centering
  \centerline{\includegraphics[width=4.8cm]{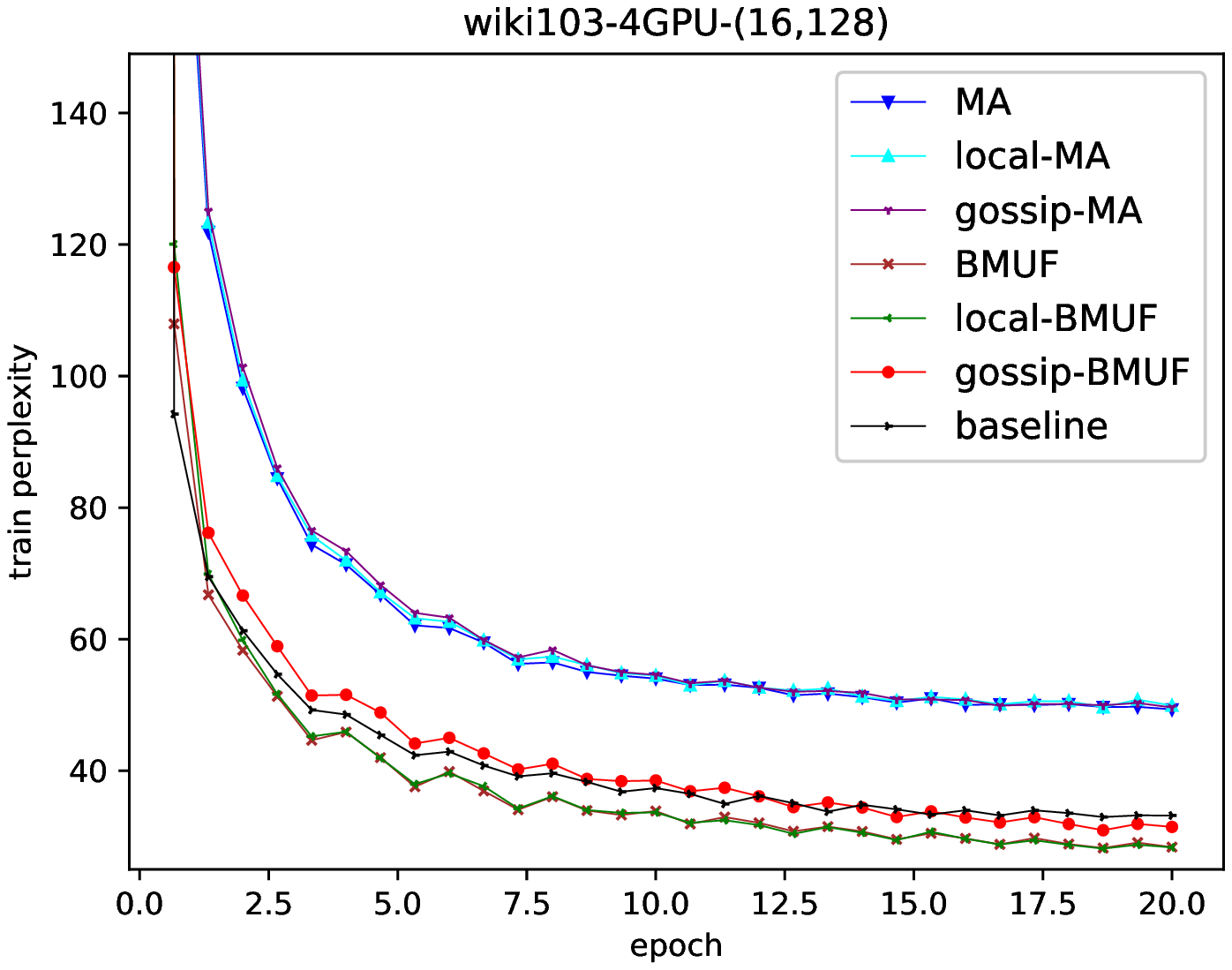}}
  %\vspace{1.5cm}
  \centerline{(b) sync-period (16,128)}\medskip
\end{minipage}
\hfill
\caption{Training curves for wiki-text-103 dataset}
\label{fig:wiki103}
\end{figure}

\begin{table}[th]
   \centering
    \setlength{\tabcolsep}{1.0mm}
    \resizebox{0.47\textwidth}{21mm}{
    \begin{tabular}{|c|c|c|c| c|c |}
      \hline
       method-period & GPUs & test ppl & period & GPUs & test ppl\\
      \hline
       MA-(8) & 4 & 55.1 & (16,128) & 8 & 64.1 \\  
       \hline
       BMUF-NBM-(8) & 4 & 50.7 & (16,128) & 8 & 50.3\\  
      \hline
       local-BMUF-(8) & 4 &  51.0  & (16,128) & 8 & 50.4 \\
     \hline
      gossip-BMUF-(8) & 4 & $\mathbf{48.4}$ & (16,128) & 8 & $\mathbf{49.0}$  \\
      \hline
       MA-(16,128) & 4 & 54.3   & (16,128) & 16 & 80.5 \\
     \hline
      BMUF-NBM-(16,128) & 4 & 51.1 & (16,128) &  16 & 50.5 \\
      \hline
      local-BMUF-(16,128) & 4 & 51.5  & (16,128) & 16 & 50.6 \\
      \hline
      gossip-BMUF-(16,128) & 4 &  $\mathbf{48.5}$  & (16,128) & 16 & $\mathbf{49.3}$\\
     \hline
     single-GPU baseline & \multicolumn{5}{|c|}{49.3}\\
     \hline
    \end{tabular}}
\caption{wiki-text-103 results}
\label{wiki-results}
\end{table}

\begin{table}[th]
   \centering
    \setlength{\tabcolsep}{1.0mm}
    \begin{tabular}{|c|c|c|}
      \hline
       method-period & GPUs & test ppl  \\
      \hline
       MA-(16,128)& 4/8/16 & 158.0/178.9/211.1 \\
      \hline
      gossip-MA-(16,128) & 4/8/16 & 158.4/180.2/214.7 \\ 
       \hline
       BMUF-NBM-(16,128) & 4/8/16 & 156.3/152.6/146.3  \\  
      \hline
       local-BMUF-(16,128) & 4/8/16 & 156.1/153.8/148.0 \\
     \hline
       gossip-BMUF-(16,128) & 4/8/16 &  $\mathbf{149.4}/\mathbf{148.4}/\mathbf{146.1}$  \\
     \hline
       single-GPU baseline  & \multicolumn{2}{|c|}{144.6} \\
    \hline
    \end{tabular}
\caption{results of Gutenberg}
\label{gutenberg-results}
\end{table}

As shown in tables \ref{wiki-results} and \ref{gutenberg-results}, gossip-BMUF consistently outperforms other methods, and even achieves a performance better than single-GPU baseline. The results of BMUF-NBM fluctuate more fiercely than that of gossip-BMUF. The performance degradation of MA is very significant when the number of GPUs is large. Local-BMUF has a slightly worse performance than BMUF-NBM, and this indicates that the randomly selected neighbors are the key success in gossip-BMUF. The training curves in figure \ref{fig:wiki103} indicate that gossip-BMUF has a very similar training performance as the single-GPU baseline. The curves of BMUF and local-BMUF implies that the over-fitting might already happen during training. We do not report the results of \textit{local-MA} and \textit{gossip-MA}, since they both have a similar performance as MA. 

Gossip-BMUF achieves speedups of $3.03$X on $4$ GPUs, and $4.95$X on $8$ GPUs, while BMUF achieve speedups of $3.20$X and $5.47$X on $4$ and $8$ GPUs respectively on wiki-text-103. The slightly degraded speed is mainly caused by the random sampling process. 

\section{CONCLUSIONS AND DISCUSSIONS}
In this paper, we present a gossip-BMUF approach to scale conventional deep learning methods to handle large scale datasets. In this approach, the network is formed with a decentralized topology, and the performance are better than conventional centralized approaches with better accuracy and lower variance. In our future work, we would like to investigate the following two directions:  (1) Evaluate our approach to other types of deep machines such as transformers \cite{vaswani2017attention}, CNNs and etc. (2) Analyze the statistical performance of \textit{gossip-BMUF}.
\bibliographystyle{IEEEbib}
\bibliography{strings,refs}

\end{document}